\documentclass[lettersize,journal]{IEEEtran}
\usepackage{amsmath,amsfonts}
\usepackage{algorithmic}
\usepackage{algorithm}
\usepackage{array}
\usepackage[caption=false,font=normalsize,labelfont=sf,textfont=sf]{subfig}
\usepackage{textcomp}
\usepackage{stfloats}
\usepackage{url}
\usepackage{verbatim}
\usepackage{graphicx}
\usepackage{cite}
\usepackage{siunitx}
\usepackage{verbatim}
\usepackage{xcolor}

\begin{document}

\title{Critical Anatomy-Preserving \& Terrain-Augmenting Navigation (CAPTAiN): Application to Laminectomy Surgical Education}

\author{
  \IEEEauthorblockN{Jonathan Wang\textsuperscript{1*},
  Hisashi Ishida\textsuperscript{2*},
  David Usevitch\textsuperscript{3},
  Kesavan Venkatesh\textsuperscript{4}, \\
  Yi Wang\textsuperscript{2},
  Mehran Armand\textsuperscript{2,5,6},
  Rachel Bronheim\textsuperscript{6},
  Amit Jain\textsuperscript{6},
  Adnan Munawar\textsuperscript{2\dag}}

  \vspace{0.5em}
  
  \IEEEauthorblockA{
  \textsuperscript{1}School of Medicine, Johns Hopkins University, Baltimore, MD, USA\\
  \textsuperscript{2}Department of Computer Science, Johns Hopkins University, Baltimore, MD, USA\\
  \textsuperscript{3}Laboratory for Computational Sensing \& Robotics, Johns Hopkins University, Baltimore, MD, USA\\
  \textsuperscript{4}School of Medicine, Vanderbilt University, Nashville, TN, USA\\
  \textsuperscript{5}Department of Mechanical Engineering, University of Arkansas, Fayetteville, AR, USA\\
  \textsuperscript{6}Department of Orthopaedic Surgery, Johns Hopkins University, Baltimore, MD, USA}

  \vspace{0.5em}
  
  \IEEEauthorblockA{
  \textsuperscript{*}These authors contributed equally to this work\\
  \textsuperscript{\dag}Corresponding author email: \texttt{amunawa2@jh.edu}}
}

\maketitle

\begin{figure*}[ht!]
    \centering
    \includegraphics[width=1\textwidth]{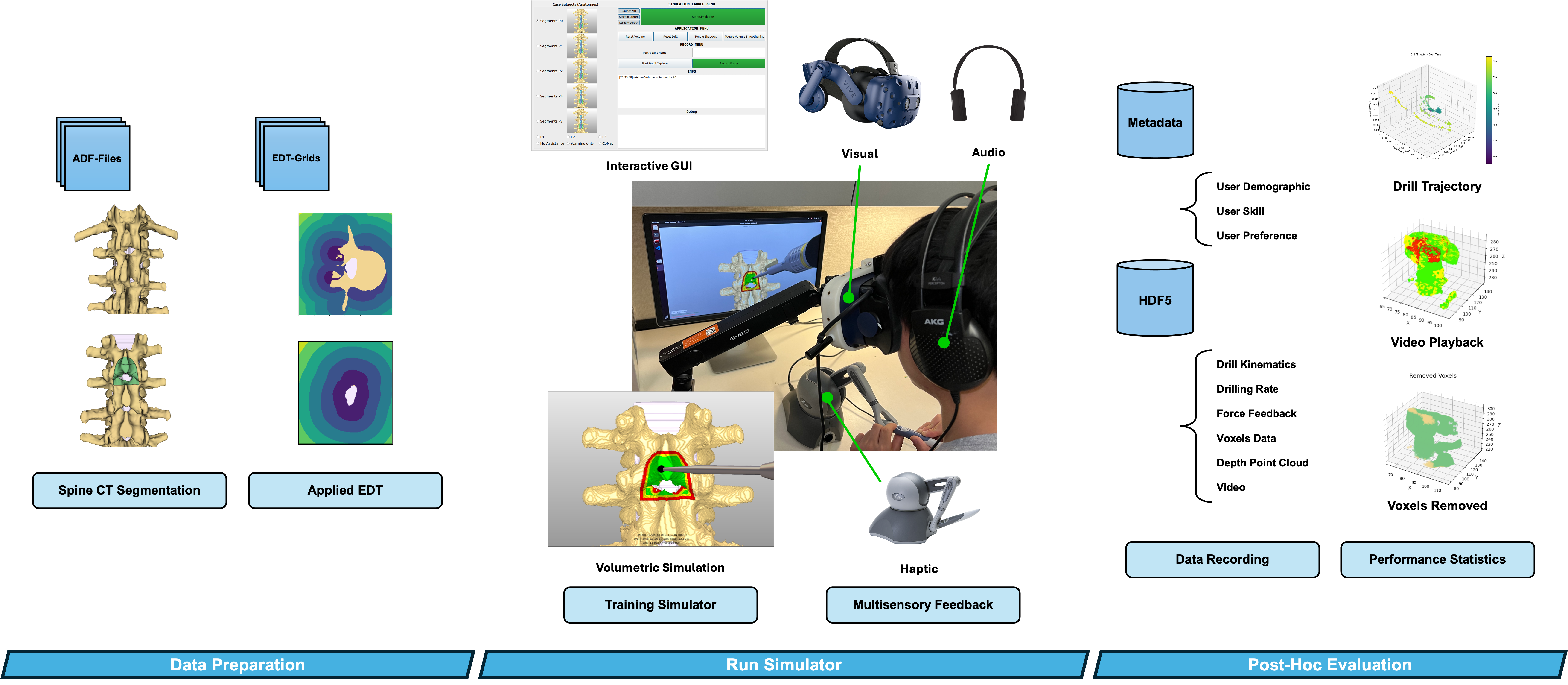}
    \caption{Overview of the immersive laminectomy simulator training pipeline. The lumbar spine was initially scanned using CT imaging and segmented in 3D Slicer to define the target regions for bone removal. Signed Distance Fields (SDF) were then applied to the segmented spine, generating the Red, Yellow, and Green boundaries. These boundaries were used to compute Euclidean Distance Transformation (EDT) Grids, which were imported into the AMBF simulator. Users can utilize the interactive GUI to select specific lumbar spines and toggle between receiving navigational assistance or not. The immersive experience was delivered to users through visual feedback via the HTC Vive Pro Headset, haptic feedback via the Phantom Omni, and auditory feedback through headphones. Additionally, the simulator records multimodal data which can be used to track performance metrics including visualizations of removed voxels, drilling time, force exerted on simulated bone, and drilling trajectories.}
\label{fig:architecture}
\end{figure*}

\begin{abstract}
Surgical training remains a crucial milestone in modern medicine, with procedures such as laminectomy exemplifying the high risks involved. Laminectomy drilling requires precise manual control to mill bony tissue while preserving spinal segment integrity and avoiding breaches in the dura—the protective membrane surrounding the spinal cord. Despite unintended dural tears occurring in up to 11.3\% of cases, no assistive tools are currently utilized to reduce this risk. Variability in patient anatomy further complicates learning for novice surgeons.
This study introduces CAPTAiN, a critical anatomy-preserving and terrain-augmenting navigation system that provides layered, color-coded voxel guidance to enhance anatomical awareness during spinal drilling. CAPTAiN was evaluated against a standard non-navigated approach through 110 virtual laminectomies performed by 11 orthopedic residents and medical students. CAPTAiN significantly improved surgical completion rates of target anatomy (87.99\% vs. 74.42\%) and reduced cognitive load across multiple NASA-TLX domains. It also minimized performance gaps across experience levels, enabling novices to perform on par with advanced trainees. These findings highlight CAPTAiN’s potential to optimize surgical execution and support skill development across experience levels. Beyond laminectomy, it demonstrates potential for broader applications across various surgical and drilling procedures, including those in neurosurgery, otolaryngology, and other medical fields.
\end{abstract}


\section{Introduction}
Conventional surgical education has historically emphasized observational learning, apprenticeship with a more senior surgeon, or practice on cadaveric specimens as means of learning a new surgical technique. However, surgery demands technical precision, rapid decision-making, and the ability to navigate diverse patient anatomies in an environment where even minor errors may lead to severe complications \cite{Pakkasjrvi2024, Nestel2025}. As such, the current educational model does not always allow trainees to hone these advanced surgical skills in high-risk anatomic areas, such as the spine. Among surgical specialties, spinal surgery stands out due to its intricate anatomy and the critical need for accuracy \cite{Tian2020, Mensah2024}. With over 4.83 million spinal procedures performed annually worldwide, the demand for precision and safety in these surgeries has never been greater \cite{Schleer2019, LealGhezzi2016}. Laminectomy—a procedure designed to relieve spinal compression by removing portions of vertebraltissue—stands out as both a common and technically demanding operation. In the United States alone, approximately 500,000 laminectomies are performed annually, underscoring its critical role in addressing spinal disorders \cite{Li2021}. However, despite its widespread use, the procedure remains fraught with challenges, particularly in balancing the need for precise decompression of the neural elements with the imperative to avoid complications, such as durotomy \cite{ajczak2024}.

Laminectomy is generally indicated when a process causes stenosis, or narrowing, of the spinal canal, compressing the neural elements. Thus, the primary surgical goal is to alleviate pressure on the thecal sac by carefully removing bone from the vertebral lamina. However, this procedure is highly skill-dependent, with variations in technique based on the surgeon’s experience, personal preferences, and the patient’s unique anatomical features. Furthermore, the spine represents an example of anatomy where critical structures, such as the dura, lie in close proximity to the operative field. Accidental breaches of the dura, known as durotomies, occur in approximately 11.3\% of cases \cite{Jankowitz2009, Bydon2015, Ishikura2017}. Such breaches can lead to severe complications, including cerebrospinal fluid leaks, infections, and long-term neurological deficits. 

The risk of durotomy is further exacerbated by the high-speed drills commonly used to perform laminectomies. To achieve precise decompression while minimizing thermal damage and excessive force application, standard high-speed medical drills ($\sim$\SI{50000}-\SI{80000} rpm) with spherical diamond burr tips ($\sim$\SI{4} {\milli\metre} dia) are commonly used. These drills allow for efficient cutting through dense cortical bone but also introduce an increased risk of incidental durotomy due to their speed and potential for skiving if not properly controlled.

The challenges of laminectomy are not limited to avoiding durotomies. Surgeons must also navigate a delicate balance between removing enough bone to decompress neural tissue and preserving sufficient structural integrity to prevent iatrogenic spinal instability \cite{Iida1990}. This requires not only technical skill but also an understanding of the patient’s unique anatomy \cite{Smorgick2015}. Yet, the current paradigm of surgical training often falls short in preparing surgeons for these complexities. As mentioned previously, traditional methods rely heavily on apprenticeship models and cadaveric dissections, where trainees learn by observing and assisting experienced surgeons. While invaluable, this approach is inherently limited by the variability in surgical cases and the high stakes of operating on the spine \cite{Atesok2012}.

In recent years, there has been growing recognition of the need for more advanced training tools that can provide realistic, immersive, and repeatable practice environments for procedural training. Physical and mixed-reality simulators have emerged as promising solutions, offering trainees the opportunity to hone their skills in a risk-free setting \cite{Boody2017, Coelho2018, TARCHALA2019, Weiss2020}. However, as surgical techniques grow more sophisticated, the limitations of current training systems have become more apparent. Existing simulators often fall short in replicating the nuanced haptic, visual, and auditory feedback that surgeons rely on during actual procedures \cite{Kadadhekar2025, Jarry2025}. For instance, during manual drilling, surgeons must listen for changes in drill frequency to detect transitions between different bone types and sense variations in thrust force to gauge depth \cite{Xia2021, Dai2021}. These subtle cues are critical for safe and effective surgery but are difficult to replicate in conventional training systems.

To address these limitations, this work presents a novel future-facing navigation and virtual training platform suited to enhance precision and safety in laminectomy and other related hand-held drilling procedures. The Critical Anatomy-Preserving \& Terrain-Augmenting
Navigation (CAPTAiN) method (seen in Figure \ref{fig:architecture}) leverages virtual reality (VR) and haptic feedback technologies to provide surgeons with an immersive, intuitive, and highly realistic training environment. At its core, CAPTAiN employs a region-based visual navigation framework that utilizes color-coded zones to facilitate precise drilling and enhance spatial awareness. Additionally, the system incorporates force feedback and auditory cues to replicate the multisensory experience of surgery. The main contributions of this paper are summarized as follows:

\begin{itemize}
  \item A novel VR-based simulator for laminectomy training and education. The simulator can load patient-specific imaging, thereby providing realistic and specific training.
  \item Enhanced realism in surgical simulation through Signed Distance Field (SDF)-based navigation, providing more immersive spatial awareness, force feedback, and audio cues compared to previous voxels-based methods.
  \item Guided learning with CAPTAiN technology for safer surgical technique.
  \item A quantitative and qualitative analysis of the CAPTAiN method versus non-navigated guidance for laminectomy milling tasks in a human subject study involving experienced residents and medical students.
\end{itemize}

\begin{figure}[htbp]
    \centering
    \includegraphics[width=0.95\columnwidth]{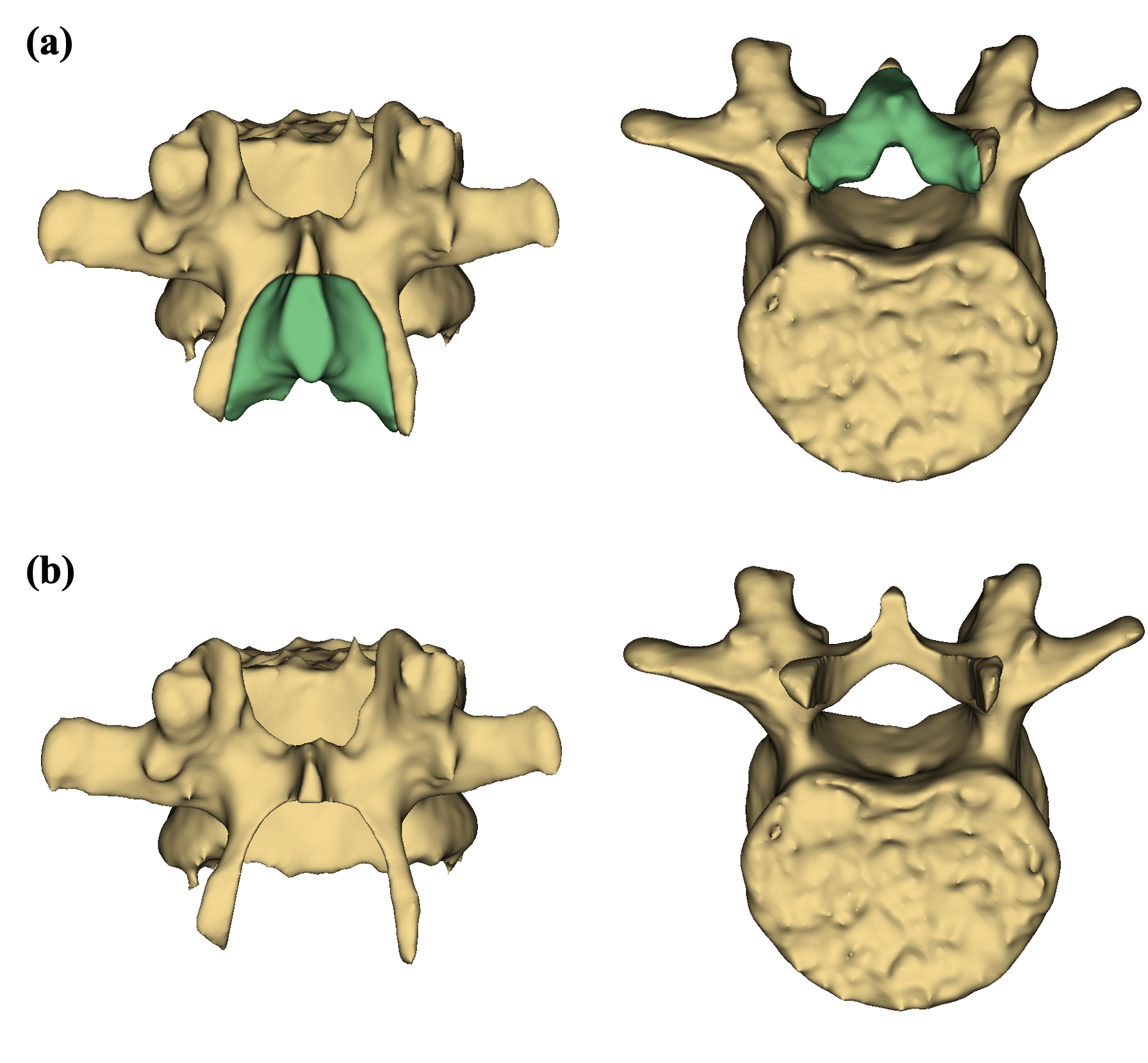}
    \caption{(a) Preoperative view of an example segmented L1 vertebra, illustrating the green-highlighted lamina region designated for removal. The visualization includes both anterior and inferior perspectives, guiding the drilling process that protects the underlying structures in the spinal canal. (b) Postoperative view displaying the remaining L1 vertebral structure following a successful laminectomy. This represents a common approach to laminectomy procedures, highlighting the structural modifications performed during the operation.}
    \label{fig:laminectomy_u_example}
\end{figure}

\section{Background}

\subsection{Anatomical Challenges: Complexity of the Spinal Lamina}
The spinal lamina’s intricate anatomy presents unique challenges for laminectomy surgery. The lamina is divided into inferior and superior halves by the pars plane, with the inferior half offering greater protection due to the presence of the ligamentum flavum—a thick, ``U''-shaped ligament that lies directly behind the inferior lamina \cite{Kim2021}. This can be visualized in Figure \ref{fig:laminectomy_u_example}. This structure serves as both a protective and easily identifiable landmark while drilling during a laminectomy, reducing the risk of dural injury. However, navigating this anatomy requires not only technical skill but also spatial intuition—an ability to mentally map bone layers and adjust drilling angles in real time.

The ligamentum flavum’s unique properties further complicate the procedure. Unlike cortical bone, which produces distinct auditory and tactile feedback during drilling, the ligamentum flavum offers less resistance, making it difficult to detect when the drill has breached the lamina. This subtlety underscores the need for training systems that replicate the sensory nuances of live surgery, enabling trainees to develop the fine motor skills and spatial awareness required for safe and effective laminectomy.

\subsection{Surgical Innovation: Templating \& Robotics}
The pursuit of safer and more efficient laminectomy surgery has driven significant innovation in both surgical techniques and training methodologies. Among these, patient-specific templating methods have emerged as a promising approach to standardize drilling paths and reduce complications. Kanawati et al. \cite{KANAWATI2021} demonstrated the efficacy of this approach, reporting no cortical breaches in 15 lumbar vertebrae procedures, with an average surgical time of just \SI{4} {\minute} and \SI{49} {\second}. These implants, designed to follow predefined drilling trajectories, minimized the risk of accidental durotomy by aligning surgical actions with patient-specific anatomies. Similarly, templating methods—widely studied in orthopedic and dental drilling tasks—have shown potential for improving precision and reducing complications \cite{Kaneyama2015, Chen2019}. However, while these methods offer notable benefits, their reliance on specialized equipment and extensive preoperative planning limits their accessibility and scalability in routine clinical settings.

Robotic systems have also made significant strides in enhancing the safety of these procedures, offering enhanced precision through a variety of control and sensing modalities. Teleoperative, shared-control, and supervisory control systems have been developed to augment surgeon capabilities, incorporating force-based, vibration-based, and image-based feedback to provide real-time guidance during drilling \cite{Wang2017, Dai2018, UsevitchDrillingReview2023}. For example, force-based systems like those developed by Wang et al. \cite{wang2010force} and Fan et al. \cite{FAN2016249} use haptic feedback to alert surgeons when the drill approaches critical structures, while vibration-based systems by Dai et al. \cite{Dai2015, Dai2021, Xia2021} detect changes in bone density to prevent over-penetration. Despite their precision, these systems often come with high costs, complex setups, and steep learning curves, making them impractical for widespread adoption. Moreover, they prioritize intraoperative support over skill acquisition, leaving trainees reliant on traditional apprenticeship models that are increasingly inadequate for modern surgical demands.

\subsection{Evolution of Surgical Training: Cadavers to Virtual Reality}
Recent advancements in simulation-based education have demonstrated significant promise in training surgeons for complex procedures \cite{Shahrezaei2024, Shah2022}. VR, augmented reality (AR), and physical simulation models offer controlled environments where trainees can refine motor skills, develop spatial awareness, and practice surgical workflows without risk to patients. These technologies have been particularly impactful in high-precision fields such as neurosurgery and orthopedics, where even minor deviations from the surgical plan can have severe consequences \cite{Polce2020, Chawla2022, Amini2024}. In laminectomy and other procedures that require bony decompression, achieving optimal bone removal while preserving neural structures is a key challenge, making effective training tools crucial for skill acquisition and patient safety. 

Traditional laminectomy training relies heavily on cadaveric dissection and direct mentorship, which, while valuable, are constrained by cost, limited availability, and ethical concerns \cite{Banaszek2017, Margalit2022}. Furthermore, the variability in cadaveric specimens and the lack of standardized feedback can hinder skill development \cite{Brenner2024}. In contrast, virtual training platforms offer an innovative and scalable alternative, enabling trainees to engage in repetitive practice with real-time guidance and objective performance assessments. A multicenter, blinded, randomized controlled trial by Lohre et al. demonstrated the effectiveness of immersive VR in surgical training. The study found that orthopedic residents who underwent VR-based training showed significantly greater improvement in technical skill acquisition compared to those trained using traditional methods \cite{Lohre2020}. These advancements in simulation-based education are reshaping the future of surgical training, offering more efficient, accessible, and effective methods to cultivate skilled, confident surgeons.

\subsection{Prior Guidance Systems for Surgical Drilling Navigation}
Parallel innovations in intraoperative navigation for other drilling procedures offer valuable insights for laminectomy training. One well-studied example is the use of virtual navigation in mastoidectomy surgery, where milling of the temporal bone is required to access the inner ear. Multiple training systems and simulators have been developed to improve accuracy and reduce the risk of complications in this procedure \cite{Hoy2017,ANDERSEN2016,Munawar2019}. CardinalSim, a VR simulator validated across Canadian Otolaryngology residency programs, is a prominent example that emphasizes general anatomical realism and procedural rehearsal \cite{Compton2020}. While procedural rehearsal is highly effective for reinforcing surgical workflows, providing real-time, adaptive feedback to guide users during drilling or signal proximity to critical anatomical structures requires novel solutions, as such feedback is uncommon in actual surgical procedures.

To address this, other systems have experimented with static visual overlays, such as displaying predefined lines or depth markers to represent drilling paths. Yet, these approaches often struggle to convey volumetric milling across complex anatomical structures. Nakao et al. \cite{Nakao2016} evaluated the use of colored-depth overlays compared to standard side navigation monitors in endoscopic drilling tasks, finding that overlay-based visualization led to shorter drilling times and improved adherence to planned drilling geometries.

Auditory guidance methods have also been integrated into surgical navigation systems. The LIVE-IGS system, developed by Dixon et al., introduced ``auditory icon'' alerts to signal proximity to critical structures, along with semi-transparent volumetric representations of sensitive anatomy \cite{Dixon2014}. Similarly, the EVADE system, created by Voormolen et al., combined color-coded navigation displays with auditory warnings to assist in temporal bone milling, alerting users when the drill approached high-risk areas \cite{Voormolen2018}. While effective, these systems were developed primarily for otolaryngology applications and did not incorporate direct overlay visualization onto the surgical field.

Even though it is plausible that novel feedback modalities could be implemented in CardinalSim, its closed-source nature limits customization and broader adoption by the research community. In contrast, the CAPTAiN system  is developed on open-source, in-house simulation platform, AMBF \cite{Munawar2019} and FIVRS \cite{munawar_fully_2023}, which provide the flexibility to implement, extend, and adapt features as needed.

\subsection{Dynamic Color-Based Guidance: A Proven Strategy}
Dynamic color-based guidance has emerged as a powerful tool for improving surgical precision. Luz et al. introduced a colored distance control method, where a virtual drill on the navigation display changed colors—green for safe regions, yellow for critical zones, and red for high-risk areas \cite{Luz2015}. Their approach demonstrated improved usability compared to a previous navigation system, reducing both task completion time and surgeon frustration \cite{Strauss2004}. Building on these principles, commercial systems like the Mako TKA system (MAKO Surgical Corp, Stryker, St. Lauderdale, FL) have applied color-coded voxel visualization for surgical guidance. In knee surgery, this system displays green voxels for target bone removal, white for successfully resected areas, and red for over-resected regions. Additionally, the drill icon changes color to indicate ``saw off'', ``approach mode'', and ``saw on'', while an audio warning is triggered if excessive velocity of instruments is detected \cite{StrykerMako}. However, while effective, these methods have been primarily used in sawing tasks rather than drilling and milling, and they rely on robotic assistance rather than fully immersive visualization.

The CAPTAiN system builds on these advancements in surgical guidance by integrating color-overlay navigation with additional multi-modal feedback, creating a training experience based on the Fully Immersive Virtual Reality System (FIVRS) for skull-base surgery \cite{munawar_fully_2023}. SDFs enable real-time distance calculations from the drill to critical structures, allowing dynamic color boundaries (green/yellow/red zones) to adapt as bone is removed. This approach, combined with patient-specific anatomies and multi-modal feedback, replicates the sensory fidelity of live surgery—from the auditory ``click'' of breaching cortical bone to the haptic resistance of ligamentum flavum. By integrating proven guidance strategies (e.g., LIVE-IGS auditory alerts) with VR immersion, CAPTAiN is a scalable, high-fidelity platform for mastering the unique challenges of laminectomy. Its focus on long-term surgical education distinguishes it from intraoperative navigation tools, offering trainees the opportunity to develop both technical skills and anatomical intuition in a risk-free environment.

\section{Methods} \label{methods}

\subsection{The CAPTAiN System}
A common surgical method for performing a lumbar laminectomy involves the surgeon milling an upside down ``U'' on the inferior half of vertebrae, along and around the laminae. This is proceeded by removing inferior bone of the spinous process and some superior portions of the adjacent vertebrae as seen in Figure \ref{fig:laminectomy_u_example}. This approach creates sufficient space to decompress the spinal cord and nerve roots while preserving the structural integrity of the remaining vertebrae. The simulation environment was designed to mimic this standard procedure, including anatomical landmarks and drilling trajectories commonly encountered in clinical practice.

CAPTAiN is a streamlined navigation approach designed to guide drilling and milling tasks across diverse spatial geometries. Many surgical navigation tasks primarily involve straight-line or path-based tool guidance, such as orienting a drill for pedicle screw placement or creating precise boreholes through bony tissue. These tasks are often carried out sequentially, where guidance depends on the current tool placement and stage of the procedure. For example, in tasks such as laminectomies, a surgeon may require guidance to initially position the drill within a specific area, then navigate around anatomical curvatures, and finally proceed with linear drilling through the target tissue. This stepwise process necessitates real-time, adaptable feedback to accommodate dynamic anatomical variations and procedural demands.

CAPTAiN facilitates user-driven volume removal by providing clear, color-coded visual feedback, enabling operators to independently determine the optimal approach for milling without relying on predefined paths. Navigation guidance is displayed simply as colored voxel regions, requiring the user to ``color in'' specified regions during drilling, as illustrated in Figure \ref{fig:ambf_screenshot_laminectomy}. Green zones represent the areas requiring removal, guiding the user to mill within these boundaries. Red zones mark critical no-drill areas designed to protect delicate structures, such as the dura or adjacent anatomical features, ensuring safe operation. Yellow zones act as warning buffers, signaling the proximity of the drilling tip to sensitive regions. These zones encourage cautionary, slower drilling to mitigate the risk of breaches. The yellow and red boundaries can be customized in thickness to suit the application. For the laminectomy cases presented in this study, these protective layers were implemented with a depth of approximately \SI{1} {\milli\metre}. In offering immediate visual feedback, the CAPTAiN system allows users to focus on precision and safety during milling while promoting a methodical and intuitive approach to navigating complex anatomical geometries.

For intuitive laminectomy navigation, the CAPTAiN system incorporates an additional rendering process within the FIVRS and Asynchronous Multi-Body Framework (AMBF) system\cite{munawar_fully_2023}. This approach dynamically applies color-coded overlays based on proximity to critical anatomy. Building on prior work utilizing SDFs to quantify distances to segmented anatomical features from computed tomography (CT) scans \cite{ishida2023improving}, CAPTAiN leverages this spatial information to adapt its graphical rendering in real time. By mapping distance-dependent visual cues onto the operative field, the system enhances spatial awareness while concurrently modulating audio and haptic feedback to heighten procedural realism. This is further described in \ref{sdfs}.

As the acting surgeon drills, colored voxels disappear corresponding to the tissue drilled, exposing previously occluded colored voxels. Upon drilling into the yellow or red regions, a visual warning text box is displayed, which is particularly helpful when the drill occludes the colors of the voxels being drilled. The volumetric scans of the lumbar spine had an approximate resolution of \SI{25959824} {\milli\metre\cubed} voxels, with each voxel covering approximately \qtyproduct{0.48 x 0.48 x 0.48}{\milli\metre} volume. For the virtual tests in this study, the bone removal rate of the drill was set to approximately 10 voxel per \SI{5} {\milli\second}. This removal rate was decreased to 1 when the drill removed the outer \SI{1.5} {\milli\metre} of the bone segment to better represent the transition of drilling between cortical (outer harder layer) to cancellous bone (inner softer bone layer) to the surgeon. A haptic stylus was used to render drilling forces, of which the average experienced by the user ranged from \qtyrange{0}{3.2}{\newton}.

\subsection{SDF Integration for Surgical Navigation \& Immersion}\label{sdfs}
\subsubsection{Calculation of SDFs}
SDFs form the core of the color-guidance and increased surgical realism found in CAPTAiN, and are computed from a 3D volume reconstructed from a segmented preoperative CT scan of the lumbar spine. The reconstruction, performed in 3D Slicer (v5.2.2), follows the methodology established by Ishida et al. \cite{ishida2023improving}. To capture the spatial context around the spinal cord, three anatomically distinct SDFs were generated. These correspond to the following regions: the vertebral foramen (VF) located posterior to the lamina, the lamina bone (LB) defining the boundary of the target area, and the sub-lamina vertebral bone (SVB) found inferior to the lamina. Each of these SDFs were computed using a 3D Euclidean Distance Transform (EDT), which determined the shortest distance from every voxel to the surface of the respective anatomical structure. Formally, for any voxel at position \( x = (x, y, z) \) in the volume, the signed distance to the surface of anatomical structure \( i \) was defined as:

\refstepcounter{equation}
\begin{equation*}
S_i(x) = EDT_i(x) =
\begin{cases} 
+\min\limits_{y \in \partial \Omega_i} \| x - y \| \text{, if } x \notin \Omega_i \\
-\min\limits_{y \in \partial \Omega_i} \| x - y \| \text{, if } x \in \Omega_i
\end{cases}
\tag{\theequation}
\end{equation*}

In this formulation, \( \Omega_i \) represents the set of voxels that constitute anatomical structure \( i \) within the 3D volume. Its boundary, \( \partial \Omega_i \), denotes the set of voxels that define the surface of the structure. A composite SDF was then constructed by assigning each voxel the minimum signed distance value across the three anatomical regions (VF, LB, and SVB). This unified field integrated spatial data from all critical regions into a single volumetric field:

\refstepcounter{equation}
\begin{equation*}
S_{\text{comp}}(x) = \min(S_{\text{VF}}(x), S_{\text{LB}}(x), S_{\text{SVB}}(x))
\tag{\theequation}
\end{equation*}

\begin{figure}[htbp]
    \centering
    \includegraphics[width=1\columnwidth]{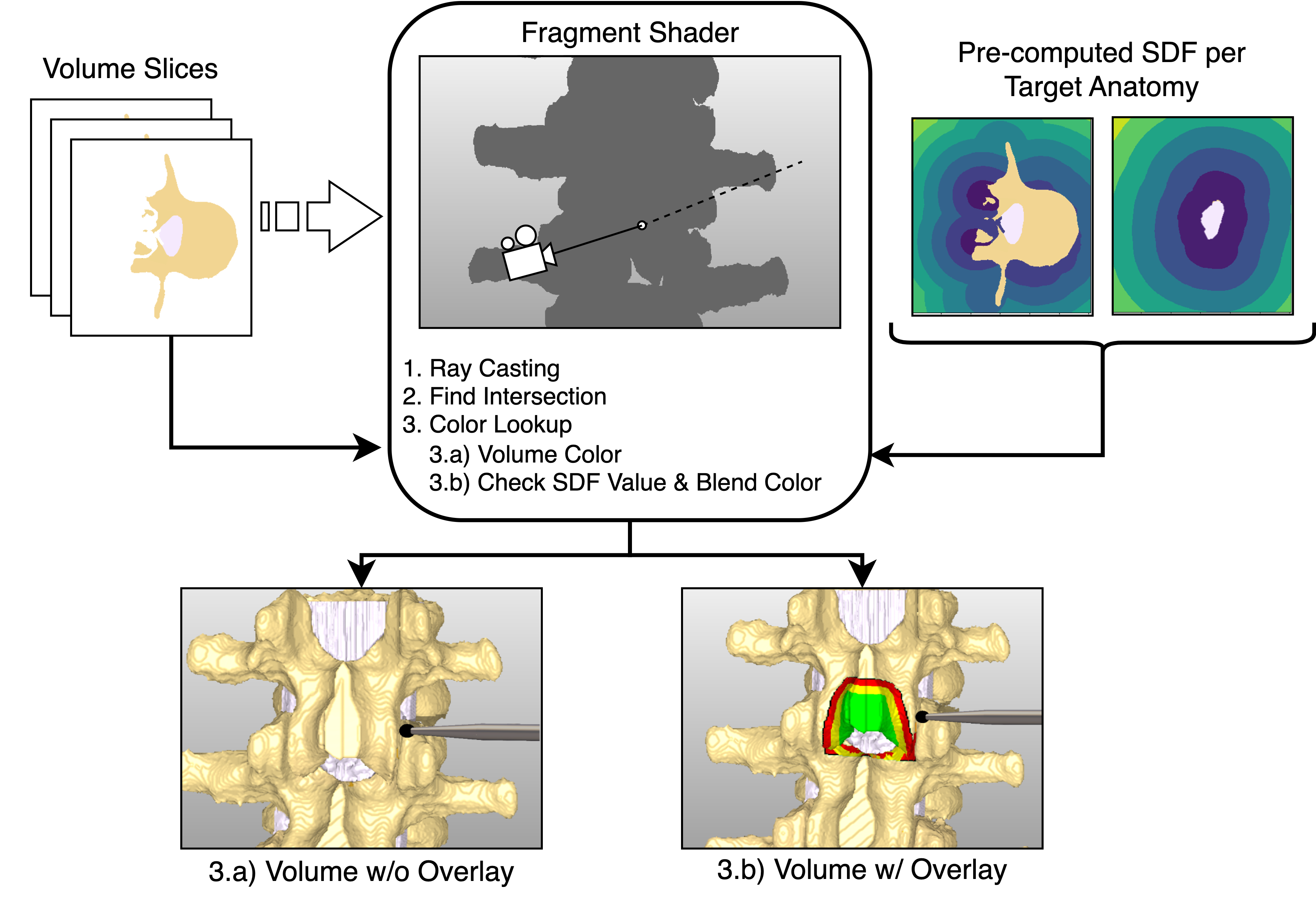}
    \caption{The system integrates preoperative CT-derived lumbar spines with precomputed Signed Distance Fields (SDFs) to provide spatially aware visual overlays. The OpenGL fragment shader dynamically queries the composite SDF texture to generate real-time, perceptually intuitive overlays, improving depth perception and situational awareness.}
    \label{fig::open_gl_shader}
\end{figure}

\subsubsection{SDF-Driven Colored Overlays}
The composite SDF was then encoded as a 3D texture and mapped to OpenGL 4.6 fragment shaders within the FIVRS/AMBF framework (Figure \ref{fig::open_gl_shader}). During real-time visualization, the shader queries the SDF texture using each fragment's 3D spatial coordinates to retrieve ($S_{\text{comp}}(x)$), which represents the minimum distance from that voxel to any of the three landmarks. Following the distance lookup, the shader utilizes the signed distance information to blend colors for each fragment in the scene. A threshold-based interpolation scheme is applied, wherein high-risk regions (those closer to critical structures) are rendered with a red gradient, moderate-risk regions with a yellow gradient, and low-risk regions with green The fragment shader employs piecewise linear blending to blend these colors seamlessly across the defined thresholds, ensuring visually continuous gradients that reflect incremental changes in proximity. This approach leverages perceptual conventions (e.g., red = danger, green = safe) to minimize cognitive load during surgical navigation.

\subsubsection{SDF-Driven Force Feedback}

To further enhance realism in haptic feedback and mimic the progressive resistance encountered in real-world drilling scenarios, CAPTAiN incorporates the laminar bone's density profiles encoded in the composite SDF ($S_{\text{comp}}(x)$). As the virtual drill interacts with the lamina, the system dynamically modulates voxel removal rates to reflect bone density gradients. Consequently, rapid excavation occurs in areas representing porous cancellous bone, while slower, resistance-laden removal reflects encounters with denser cortical layers. This density-dependent modulation is mirrored in haptic feedback through the proxy-based haptic rendering method, where the exerted force magnitude scales inversely with the removal rate \cite{Munawar2021}. The resulting tactile profile replicates the progressive ``bite'' of a surgical burr as it transitions between bone strata, complete with subtle vibrations that emulate the auditory-tactile coupling of real-world drilling. Formally, the haptic feedback force ($F_{\text{haptic}}$) is defined as:

\refstepcounter{equation}
\begin{equation*}
    F_{haptic} = k (\vec{x}_{goal} - \vec{x}_{phy}) 
\tag{\theequation}
\end{equation*}

Here, $ k \in \mathbb{R} $ is a predefined stiffness constant, and $\vec{x}_{\text{goal}}, \vec{x}_{\text{phy}} \in \mathbb{R}^6$ represent the commanded (desired) and simulated positions of the drill tip, respectively. The user's haptic device input directly determines $\vec{x}_{\text{goal}}$, while $\vec{x}_{\text{phy}}$ responds according to the simulated physical interactions with the virtual environment. As described in \cite{Munawar2021}, the simulated drill tip ($\vec{x}_{\text{phy}}$) attempts to follow the commanded position ($\vec{x}_{\text{goal}}$), but its motion is modulated by contact dynamics. Thus, in higher-density regions characterized by greater $S_{\text{comp}}(x)$ values, the simulated drilling motion is intentionally slowed, enhancing resistance and generating higher force feedback to mimic realistic tactile sensations.

\subsubsection{SDF-Driven Audio Feedback}
CAPTAiN leverages SDF-based audio modulation to enhance the realism of the drilling experience and provide intuitive auditory cues for surgical navigation. By dynamically adjusting the pitch of the drill sound based on the voxel composition at the drill tip, the system enables users to distinguish between different bone densities, such as cortical and cancellous bone, in real time.

The simulated frequency of the drill sound, \( f_{\text{drill}} \), is modulated as a function of the SDF and the material properties of the underlying bone:

\refstepcounter{equation}
\begin{equation*}
f_{\text{drill}} = f_{\text{base}} + \Delta f \cdot \alpha(S)
\tag{\theequation}
\end{equation*}

In this representation, \(f_{\text{base}}\) is the baseline drill pitch in cancellous bone, \(\Delta f\) is the frequency shift applied based on the material density, and \(\alpha(S)\) is a normalized mapping function derived from the local SDF that reflects the density and composition of the bone at the drilling site. Specifically, \(\alpha(S)\) maps SDF values to a range [0, 1], where lower values correspond to softer cancellous bone regions and higher values to denser cortical bone. Although the precise analytical form of \(\alpha(S)\)  employed in simulations is context-specific and simulation-dependent, its general purpose is to provide a continuous, smooth mapping from voxel density to auditory frequency modulation. Hence, cortical bone, being denser and more rigid, produces a higher-pitched, sharper drill sound, whereas cancellous bone generates a lower-pitched, softer sound. This audio feedback mirrors real-world surgical drilling, where the pitch of the drill changes as it transitions between different bone layers \cite{Nigam2022}.

\subsection{Participants}
Eleven orthopedic residents or medical students (8 male, 3 female) from the Johns Hopkins Orthopedic Surgery Department volunteered as test participants, and informed consent was obtained from each individual prior to participation. All the participants self-reported normal vision, auditory, and motor skills. The study was approved by the Johns Hopkins Hospital IRB committee under IRB approval number IRB00343800. None of the participants had used the hardware or software contained in the testing setup previously.

\subsection{Experimental Design}
The experiments followed a repeated-measures design to evaluate the effects of three different navigation methods on surgical performance. Key response variables measured during the experiment included the total voxels removed, total drilling time, and the number of breaches. To minimize potential order effects, the drilling scenarios were counterbalanced. Each participant completed 10 pre-randomized laminectomy test scenarios, with five laminectomies assigned to each navigation method. These scenarios were presented in a randomized order to ensure unbiased evaluation of the performance metrics across both navigational methods.

\subsection{Setup}
\subsubsection{Preoperative Planning} 

\begin{figure}[h!]
    \centering
    \includegraphics[width=1\columnwidth]{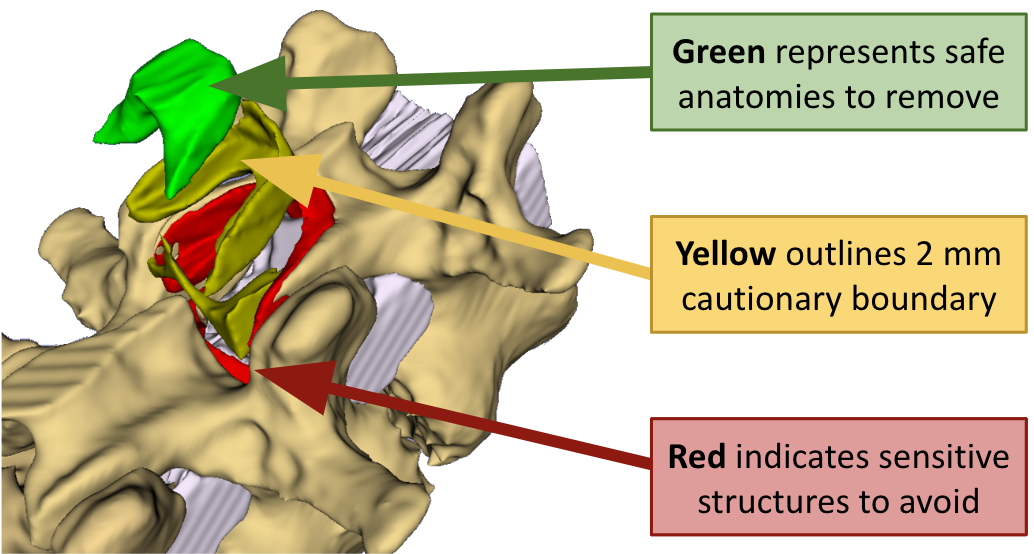}
    \caption{Example of the red, yellow, and green color schemes adopted by CAPTAiN. The green area denotes necessary drilling structures, yellow is a cautionary layer indicating proximity to dangerous structures and the need for careful drilling, and red indicates barriers not to be breached. SDF boundaries were set at \SI{1} {\milli\metre} from the vertebral foramen boundary anatomy in red and at a low \SI{0.1} {\milli\metre} boundary laterally. A \SI{1} {\milli\metre} cautionary layer in yellow was set in front of all red boundaries.}
    \label{fig:color_nav_segmentations}
\end{figure}

Five lumbar spine CT scans (male and female) were selected to reflect a representative range of anatomical variability relevant for laminectomy simulation. This sample size was chosen to capture inter-patient differences in vertebral morphology, such as lamina thickness, curvature, and foramen dimensions, which can impact surgical approach. The initial CT scans (comprising T12 and L1-L4 segments) were imported into 3D Slicer and segmented to isolate the vertebrae of interest (L1-L3) for the laminectomy procedure. Using the anatomical features of the spine, a simple upside-down ``U''-shaped resection path was designed as seen in Figure \ref{fig:laminectomy_u_example}, with adjustments made for the specific characteristics of each spine.

The segmented vertebral regions were then subdivided into five distinct segments (A, B, C, D, and E) using a combination of the ``Scissors'' and ``Logical Operators'' tool in 3D Slicer, with each subdivision representing a portion of the vertebra to be removed during the procedure. Using the ``Scissors'' tool, portions of the segmented vertebrae were selected to match the desired anatomical regions corresponding to the laminectomy path. This allowed precise separation of sections A through E, ensuring each segment accurately represented the intended resected portion of the vertebra. The ``Logical Operators'' tool was then used to incorporate additional regions, refining the boundaries and ensuring proper alignment with the anatomical features observed in the CT scans. Each of the five segments was carefully delineated to ensure that the segmentation matched the predefined ``U''-shaped resection path. 

These segments were further processed by applying SDF on the lateral edges of each section, as can be seen in Figure 
\ref{fig:color_nav_segmentations}. The SDF boundaries were set in red and yellow to demarcate the areas of drilling, with red indicating the vertebral foramen and yellow applied to the boundaries below the pars plane. The lateral edges of sections A through D were designated for green coloring, indicating areas where the drilling was intended to take place. The SDF parameters were carefully chosen to ensure protection against durotomies, especially in the regions near the dura and spinal column ``no-puncture'' zone. The red and yellow boundaries were set to a thickness of \SI{1} {\milli\metre}, while the red vertebral foramen boundary was defined at \SI{0.1} {\milli\metre} to provide a fine margin of safety for drilling procedures. This segmentation methodology, based on the anatomical considerations and SDF application, ensured accurate and safe representation of the vertebral anatomy during the planning phase of the laminectomy.

On average, the entire workflow, from segmentation to SDF application, required approximately \SI{30} {\minute} per scan. This estimate reflects processing on a standard workstation and may vary depending on the resolution of the CT data and computational performance of the platform used. The most time-intensive step involved manual segmentation of the vertebrae, which was prioritized in this study to ensure anatomical fidelity and precision in defining drillable versus critical zones.

\subsubsection{Materials}

\begin{figure}[htbp]
    \centering
    \includegraphics[width=1\columnwidth]{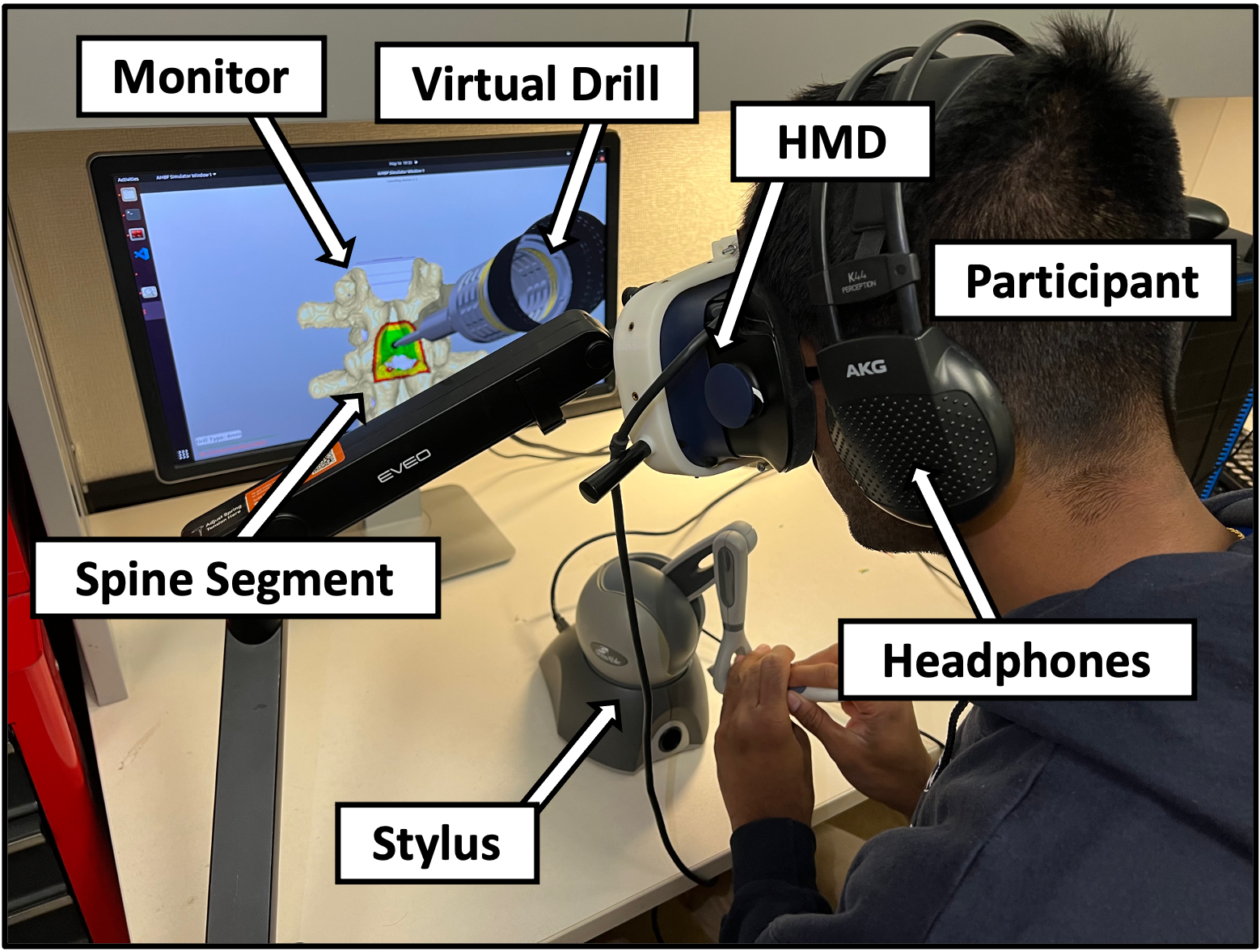}
    \caption{The human participant testing took place at a desk shown with a mock surgical microscope setup which used an HTC Vive Pro HMD for viewing the environment. Audio was played through the headphones, and a Geomagic Touch stylus was used for manipulating the virtual drill. The 3D view on the monitor was visualized in the VR HMD. Not depicted is the 2D image of the spine with the completed cut used as a guide for both the non-guided navigated and CAPTAiN methods. The ``participant'' seen here was not one of the participants of the study.}
    \label{fig:hardware_setup}
\end{figure}

The physical setup for the experiment can be seen in Figure \ref{fig:hardware_setup}. An HTC Vive Pro HMD display was mounted to an adjustable table-mounted arm via a 3D printed fixture. The mounted display acted as mock surgical microscope and was ergonomically adjustable for each participant. A desktop computer with AMD Ryzen 7 5800 CPU, 32 GB DDR4 RAM, and a NVidia RTX 3080 GPU interfaced the VR HMD for visual output, and headphones for audio output that included drilling sound, and warning sounds from the CAPTAiN system. 

Preoperative spine segments were imported from 3D Slicer, and the virtual drill was controlled with a Geomagic Touch haptic stylus, as shown in Figure \ref{fig:architecture}. The stylus provided force feedback and allowed users to adjust their view of the spine segments. Headphones played a drilling sound that changed with drilling depth. A footpedal was used to turn the drill on and off allowing the user to use the drill tip to touch and feel (probe) anatomy when not drilling if desired. 

\begin{figure}[htbp]
    \centering
    \includegraphics[width=1\columnwidth]{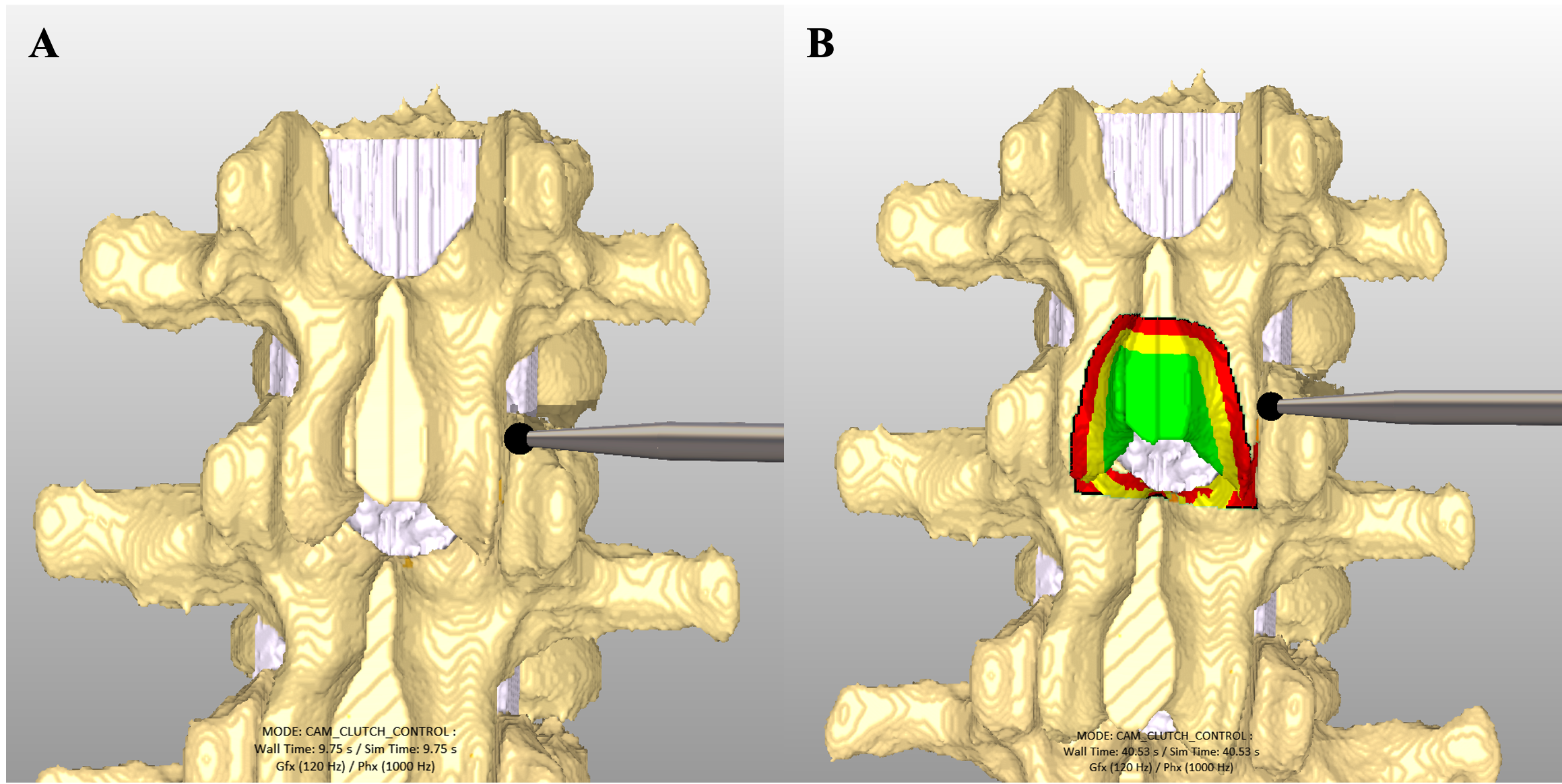}
    \caption{(a) Simulator view of a laminectomy procedure without CAPTAiN, where voxel colors are not displayed to guide the user. (b) Simulator view with CAPTAiN enabled, highlighting voxel regions using a color-coded system (green, yellow, and red) to indicate surgical guidance for safe and accurate resection. The addition of color-based navigation provides visual cues to assist users in identifying the target and critical regions during the procedure. Users were tasked with drilling green and yellow zones while avoiding red zones or other bone colored normal anatomy.}
    \label{fig:ambf_screenshot_laminectomy}
\end{figure}

\subsection{Procedure} \label{procedure}

Participants were seated in a comfortable chair at a desk and were encouraged to adjust the chair, headset, and stylus positions to their ergonomic preferences. Each participant performed two practice laminectomy drilling tests to familiarize themselves with the virtual drilling system. The first test employed the non-navigated method on the L2 vertebra, while the second utilized the CAPTAiN method on the L3 vertebra. Visual differences between these two navigation techniques are illustrated in Figure \ref{fig:ambf_screenshot_laminectomy}. During each test, participants were tasked with completing a single laminectomy as previously described in \ref{procedure}. To ensure comfort and adaptation to the system, participants were allowed unlimited time to complete these practice sessions, with no performance metrics being recorded during this phase. This approach ensured that participants could focus on becoming accustomed to the navigation and control of the drill within the virtual environment. Participants were instructed to hold the stylus, which simulated the drill handle, in a two-handed grip consistent with standard laminectomy drilling techniques \cite{Langeveld2018}. This orientation provided a realistic simulation of surgical practices, enabling participants to develop proficiency with the system before proceeding to the experimental phase of the study.

A set of 12 spine segments (L1-L3) was created,  with five segments per group randomly assigned to either the non-navigated or CAPTAiN navigation condition, while the remaining two were used for practice sessions to introduce each navigation method. For both the non-navigated and CAPTAiN tasks, a 2D reference image of the spine segment with a complete laminectomy was displayed without any colored guidance. This reference served as a visual guide for the participants during the tasks. Following the two practice cases designed to familiarize participants with the equipment and protocols, participants performed laminectomy drilling tests on the 10 pre-randomized spine segments. The test order was randomized to minimize potential order effects. To prevent fatigue, participants were required to take a break of at least one minute following the practice session and after every four laminectomies. Participants then completed three surveys: the NASA Task Load Index (NASA-TLX) survey \cite{HART1988139} to assess their workload and experience with the two navigation methods, a demographics survey to capture prior experience with drilling tasks, and a qualitative feedback survey to evaluate their perceptions of the CAPTAiN system.

\subsection{Performance Metrics}
For each of the 10 laminectomy drilling tasks, voxel-level data was recorded in real-time, including the timestamp, color, and spatial position of each voxel removed by the drill tip. Post-hoc analysis was performed to determine key performance metrics, including surgical completion rates, total drilling time, and the number of breaches. Surgical completion rates were calculated as the percentage of primary (green), secondary (yellow), and critical (red) voxels removed compared to the guidance zones, providing an assessment of procedural accuracy and adherence to the desired resection boundaries. Total drilling time was measured from the timestamp of the first voxel removed to the timestamp of the last voxel removed for each laminectomy task, representing the duration of each surgical simulation.

Breaches were quantified as the number of instances where the drill tip entered critical (red) zones or actual anatomy continuously for a minimum threshold of 5 voxels. To ensure breaches were not overcounted, an additional temporal threshold of at least \SI{2} {\second} was applied, requiring consecutive voxel breaches within this timeframe to be counted as a single event. This approach enabled a robust measurement of breaches into any critical zone.

Additional metrics included subjective workload assessments captured through NASA-TLX post-survey scores. Participant demographics, including gender (male/female), years of experience performing laminectomies, and the frequency of drilling tasks in their clinical practice, were also collected. The latter was categorized as either 0 (``Lack of Drilling'') or 1 (``Regular Drilling''). These factors were included to assess their potential impact on performance outcomes, providing a multidimensional evaluation of surgical proficiency under experimental conditions.

\subsection{Statistical Analysis}
Statistical analyses were conducted to evaluate the performance metrics and subjective workload data collected during the study. One-sided paired t-tests were performed were performed on the qualitative post-hoc analyzed performance metrics, including total drilling time, surgical completion rates, and the number of breaches, with the alternative hypothesis positing that the CAPTAiN system would outperform the non-navigated method in each case.

To further explore the impact of participant demographics on performance outcomes, results for surgical completion rates were stratified based on the frequency of drilling tasks in participants’ clinical practice and their years of prior experience performing laminectomies. One-sided paired t-tests were applied to the performance metrics, providing a robust non-parametric framework to assess differences between the stratified groups in relation to the navigation method.

Similarly, the NASA-TLX survey data were analyzed to assess cognitive workload. A one-sided paired t-test was performed to compare workload ratings between the CAPTAiN and non-navigated methods, with the hypothesis that CAPTAiN would reduce cognitive load and improve user experience relative to the non-navigated approach. All statistical analyses were conducted with a significance threshold of $p\leq0.05$.

\section{Results}
\subsection{Comparison of Drilling Guidance Methods}
\begin{figure}[htbp]
    \centering
    \includegraphics[width=1\columnwidth]{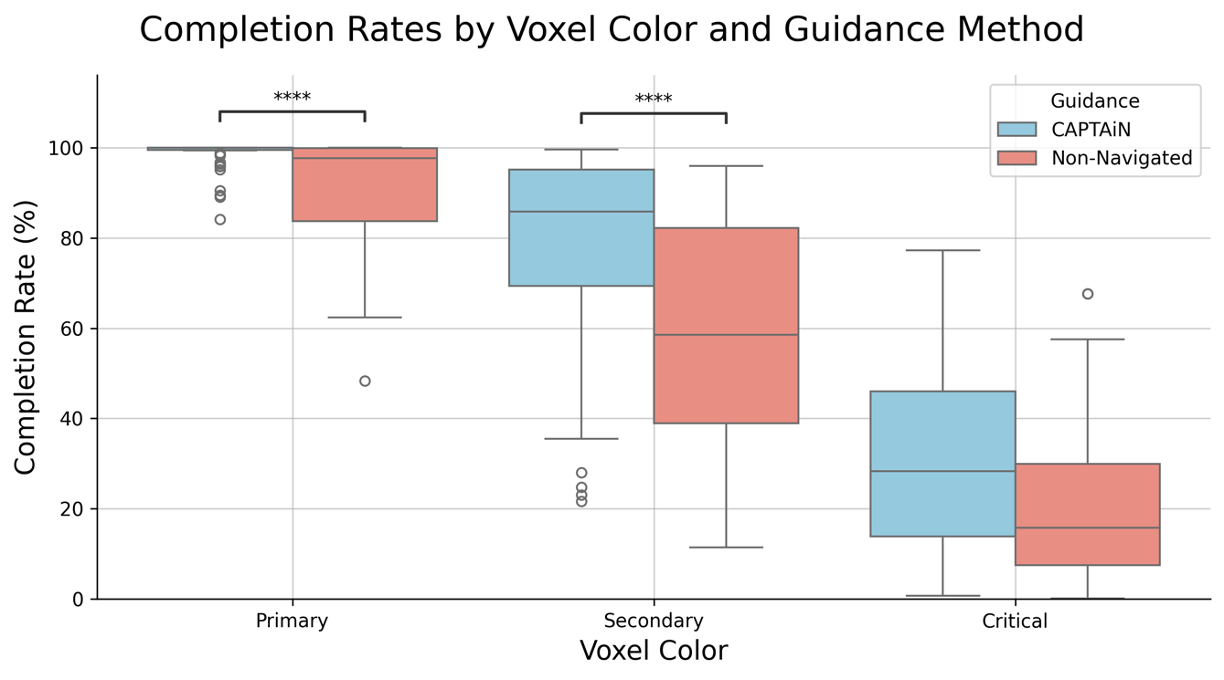}
    \caption{Grouped box plots illustrate the completion rates (\%) for primary (green voxels), secondary (yellow voxels), and critical (red voxels) anatomical targets under the CAPTAiN (blue) and non-navigated (red) guidance methods, averaged across all participants. Results indicate that the CAPTAiN method significantly improved completion rates for primary and secondary anatomical targets compared to non-navigated. No statistically significant difference was observed in the completion rates for critical anatomical targets. Statistically significant differences between guidance methods are indicated by $p\leq0.0001$ (****).}
    \label{fig:color_comp_rates}
\end{figure}

The CAPTAiN guidance method demonstrated significant improvements in surgical precision and efficacy compared to the non-navigated approach, as seen in Figure \ref{fig:color_comp_rates}. Specifically, the average completion rates of green voxels, representing the highest priority surgical targets, were significantly higher under CAPTAiN (98.66\% vs. 90.76\%, $p=9.85\times10^{-6}$). Similarly, yellow voxels, which represent secondary targets, also exhibited significantly higher completion rates with CAPTAiN (77.48\% vs. 58.47\%, $p=1.74\times10^{-7}$). Importantly, the enhanced performance in targeting primary (green) and secondary (yellow) voxels did not lead to increased removal of critical (red) voxels to avoid. The average completion rates of critical voxels were not significantly different between the two methods (30.79\% vs. 20.46\%, $p=0.99$), demonstrating that CAPTAiN allowed for greater precision without compromising safety by overextending into undesired areas.

Furthermore, CAPTAiN did not result in a significant increase in operative duration or risk. Total drilling time under CAPTAiN was comparable to the non-navigated approach (\SI{242} {\second} vs. \SI{220} {\second}, $p=0.25$), and the average total number of breaches into the critical (red) region was not statistically significant (17 vs. 17, $p=0.83$).

\subsection{Stratification Based on Prior Experience}
\begin{figure}[htbp]
    \centering    \includegraphics[width=1\columnwidth]{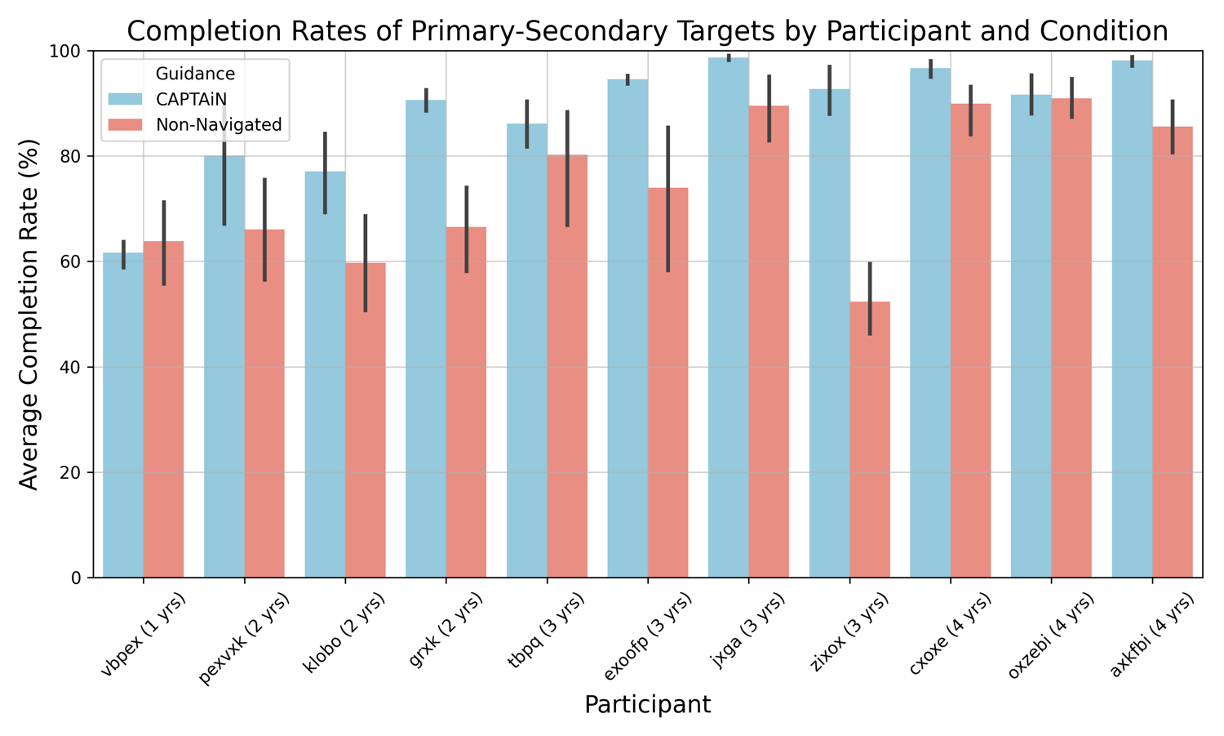}
    \caption{Average completion rates (\%) of primary and secondary anatomical targets (green and yellow voxels) by participant under CAPTAiN (blue) and non-navigated (red) guidance conditions. Higher completion rates indicate more optimal voxel removal during simulated laminectomy. Participants are ordered by self-reported years of laminectomy experience, showing that as experience increases, the gap in completion rates between CAPTAiN and non-navigated conditions narrows. This suggests reduced reliance on guidance systems for more experienced users.}
    \label{fig:comp_rates_experience}
\end{figure}

To assess the impact of participant experience, surgical completion rates for primary (green) and secondary (yellow) voxels were further stratified based on self-reported drilling frequency in their medical practice (``Regular Drilling'', n = 6; ``Lack of Drilling'', n = 5) and years of prior experience performing laminectomies (1 year, n = 1; 2 years, n = 3; 3 years, n = 4; 4 years, n = 3). CAPTAiN demonstrated superior primary voxel completion rates compared to the non-navigated method for all but the most experienced cohort with 4 years of prior experience. Participants with 1-3 years of experience showed significant increases under CAPTAiN: 1-year (96.32\% vs. 87.06\%, $p=0.023$), 2-year (97.59\% vs. 86.40\%, $p=0.0024$), and 3-year (99.90\% vs. 89.10\%, $p=0.0025$) groups. Notably, the 4-year cohort exhibited near-equivalent completion rates for primary voxels between CAPTAiN (98.86\%) and non-navigated (98.58\%). For secondary voxels, CAPTAiN maintained improved performance across all groups except for the 1-year participant, where the non-navigated method (41.53\%) led to higher completion rates compared to CAPTAiN (29.22\%). Significant increases in secondary voxel completion rates were observed for 2-year (67.77\% vs. 42.37\%, $p=0.00069$), 3-year (86.08\% vs. 59.16\%, $p=6.72\times10^{-5}$), and 4-year (91.79\% vs. 79.28\%, $p=0.0027$) cohorts. 

\begin{figure*}[htbp]
    \centering
    \includegraphics[width=1\textwidth]{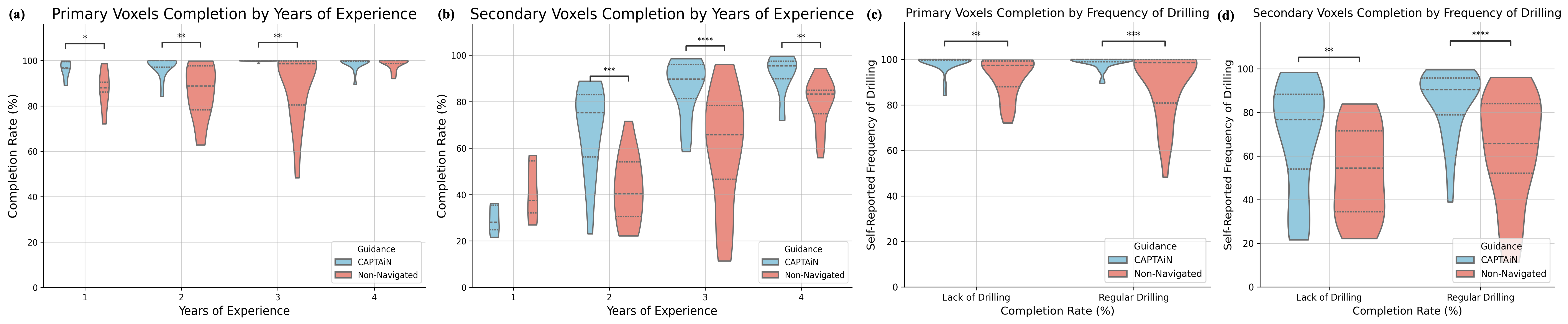}
    \caption{Grouped violin plots compare completion rates (\%) of (a) primary anatomical targets (green voxels) and (b) secondary anatomical targets (yellow voxels) stratified by participants’ self-reported laminectomy experience (years), and (c) primary and (d) secondary anatomical targets removal categorized by self-reported drilling frequency. Top panels (a-b): Participants with 1, 2, and 3 years of experience demonstrated significantly higher completion rates of primary anatomical targets using CAPTAiN (blue) versus non-navigated methods (red). For secondary anatomical targets, CAPTAiN significantly improved completion rates across 2–4 years of experience, though the performance gap narrowed with increasing expertise. Bottom panels (c-d): CAPTAiN significantly enhanced completion of primary and secondary anatomical targets for inexperienced drillers (``No Regular Drilling'' group). The same can be noted for the enhanced completion of primary and secondary anatomical targets for frequent drillers (``Regular Drilling'' group). Median completion rates increased universally with CAPTAiN, irrespective of drilling familiarity. Statistically significant differences between guidance methods are indicated by $p\leq 0.05$ (*), $p\leq 0.01$ (**), $p\leq0.001$ (***), and $p\leq0.0001$ (****).}
\label{fig:voxels_by_exp_freq}
\end{figure*}

The comparative performance of all participants across years of prior experience, including the aggregated completion rates for both primary (green) and secondary (yellow) voxels under both CAPTAiN and non-navigated conditions, is visualized in Figure \ref{fig:comp_rates_experience}. As the number of years of experience increased, the difference in completion rates between the two guidance methods diminished, with the most minimal difference observed for participants with 4 years of experience, where the completion rates of primary voxels were nearly identical between CAPTAiN and non-navigated. Differences in the completion rates of secondary voxels also seemingly followed this trend, which can be noted in Figure \ref{fig:voxels_by_exp_freq}. 

In all groups stratified on self-reported drilling frequency in medical practice, the CAPTAiN method consistently outperformed the non-navigated method. Among participants reporting ``Lack of Drilling'', CAPTAiN achieved significantly higher average completion rates for both primary (green) voxels (98.58\% vs. 92.51\%, $p=0.0011$) and secondary (yellow) voxels (68.50\% vs. 52.95\%, $p=0.0032$). Similarly, in those reporting ``Regular Drilling'', CAPTAiN maintained an increase, with average completion rates for primary voxels (98.73\% vs. 89.31\%, $p=0.00083$) and secondary voxels (84.96\% vs. 63.07\%, $p=5.96\times10^{-6}$) markedly increased. As observed in the stratification by years of prior experience, the difference in completion rates between CAPTAiN and non-navigated was more pronounced for participants with less frequent drilling. For participants reporting ``Regular Drilling'', the performance gap in completion rates for primary and secondary voxels decreased, indicating that CAPTAiN's impact is more pronounced in those with less frequent drilling experience.

\subsection{Post-Completion NASA-TLX Survey}

\begin{figure}[htbp]
    \centering
    \includegraphics[width=1\columnwidth]{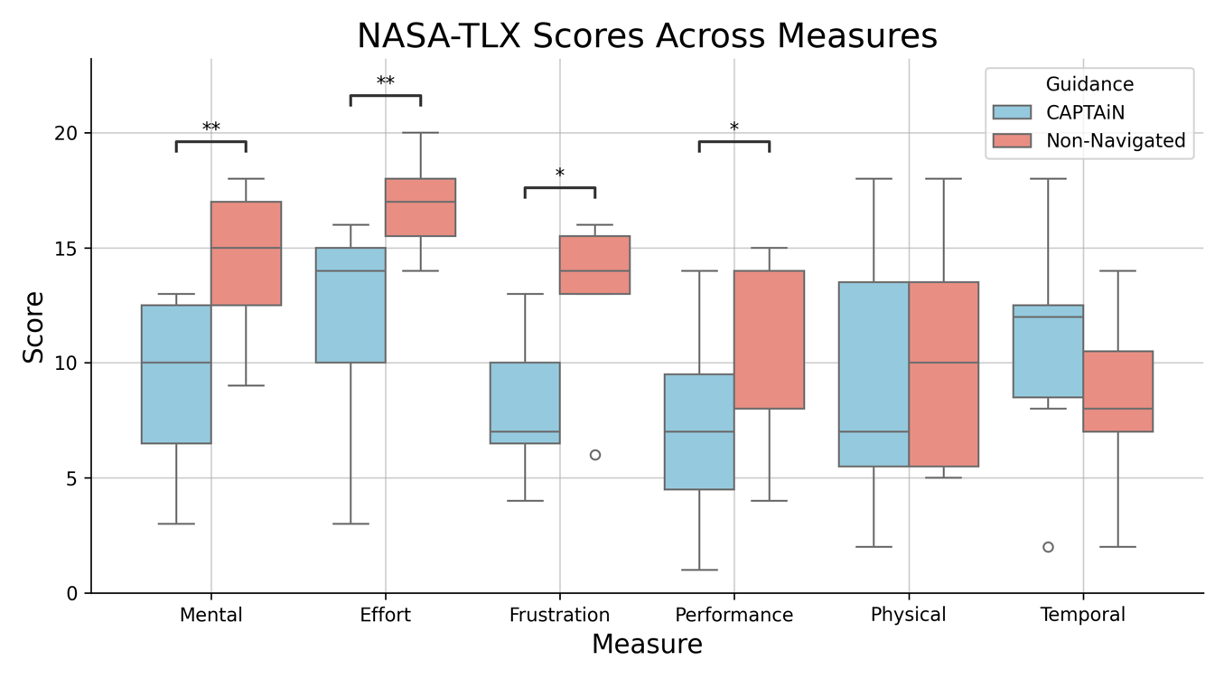}
    \caption{NASA-TLX survey results. The box plot compares self-reported NASA-TLX workload scores across six measures (Mental, Effort, Frustration, Performance, Physical, Temporal) under CAPTAiN (blue) and non-navigated (red). Data were collected from all 11 participants. Statistically significant differences were identified in the Mental, Effort, Frustration, and Performance categories, with the CAPTAiN condition resulting in lower perceived workload. CAPTAiN also outperformed non-navigated for all areas except for the Temporal category. Statistical significance is denoted as $p\leq0.05$ (*) and $p\leq0.01$ (**).}
\label{fig:nasa_tlx}
\end{figure}

Survey results from the NASA-TLX assessment were collected from all 11 participants (see Figure \ref{fig:nasa_tlx}), with participants rating their experience across six dimensions: Mental Demand, Effort, Frustration, Performance, Physical Demand, and Temporal Demand. Ratings ranged from 1 to 20, with lower numbers representing lower workload ratings. In this assessment, CAPTAiN outperformed non-navigated in almost every metric. Specifically, CAPTAiN significantly reduced the ratings for Mental Demand (9.14 vs. 14.43, $p=0.0025$), Required Effort (11.86 vs. 16.86, $p=0.0074$), Frustration (8.14 vs. 12.86, $p=0.0115$), and Performance (7.14 vs. 10.14, $p=0.0499$). These findings indicate that using CAPTAiN is more intuitive, requires less effort, results in less frustration overall, and leads to better performance outcomes.

\section{Discussion}

This study aimed to assess the impact of the CAPTAiN system compared to traditional non-navigated guidance in the context of simulated laminectomy procedures. The findings suggest that CAPTAiN significantly improves the completion rates for primary (green) and secondary (yellow) voxels without compromising safety or operative duration. These results provide insights into the potential for CAPTAiN to enhance surgical precision, particularly for less experienced users.

\subsection{Enhancement of Surgical Precision with CAPTAiN}

The results presented in Figure \ref{fig:color_comp_rates} demonstrate that CAPTAiN-assisted procedures achieved significantly higher completion rates for both primary (green) and secondary (yellow) target voxels compared to the non-navigated approach. This improvement is consistent with the primary objective of CAPTAiN: to enhance targeting accuracy and reduce the risk of inadvertent damage to critical regions. Furthermore, the comparable improvement in the completion rates for secondary voxels under CAPTAiN emphasizes its effectiveness for maintaining precision across ancillary surgical targets.

One of the critical concerns when implementing any new surgical technology is its impact on operative time. Fortunately, the use of CAPTAiN did not significantly increase the total drilling time compared to non-navigated methods. This finding suggests that the integration of CAPTAiN into the surgical workflow may not negatively affect the duration of the procedure, which is important for maintaining efficiency in clinical practice. Furthermore, the number of breaches into the critical (red) region was equivalent between the two methods, reinforcing that the enhanced precision provided by CAPTAiN does not come at the cost of safety or increased risk to the patient.

\subsection{CAPTAiN's Impact on Varying Experience Levels}

The stratified analysis of user experience levels reveals important insights into CAPTAiN’s role in surgical training. Notably, the system demonstrates differential benefits across varying levels of experience, as detailed in Figure \ref{fig:voxels_by_exp_freq}. For less experienced users (e.g., those with 2 years of experience or less frequent exposure to drilling), CAPTAiN significantly improved both primary (green) and secondary (yellow) voxel completion rates, suggesting it compensates for deficits in spatial awareness and precision common in early-stage practitioners. Among more experienced users (e.g., those with 4 years of experience), CAPTAiN's impact was more nuanced. While primary voxel completion rates were nearly identical between CAPTAiN and non-navigated methods, secondary voxel performance remained significantly enhanced under CAPTAiN. This indicates that while those more well-versed in laminectomies or drilling are able to independently achieve a baseline proficiency, CAPTAiN still optimizes performance in more complex task components that require refined precision, ultimately enhancing accuracy and reducing variability.

The consistent improvement in secondary (yellow) voxel outcomes across all experience levels suggests that the real-time navigational feedback from CAPTAiN provides universal value, particularly in more technically demanding aspects of the procedure. However, the diminishing performance gap for primary (green) voxels among those more experienced implies that fundamental surgical navigation becomes internalized with experience, reducing reliance on external guidance systems. These findings position CAPTAiN as a tool with dual utility: for novices, it serves as a compensatory aid to bridge experience gaps, while for more seasoned practitioners, it functions as a precision-enhancing safeguard against variability. This adaptability underscores its potential to standardize surgical outcomes across heterogeneous operator skill levels.

\subsection{Cognitive Load and Surgeon Fatigue}

Participants consistently reported lower scores across all NASA-TLX domains, except for Temporal demand, when using CAPTAiN, indicating a reduction in cognitive strain often associated with complex and high-stakes surgical tasks (Figure \ref{fig:nasa_tlx}). Notably, the significant reductions in Frustration, Effort, Mental, and Performance indicate that CAPTAiN not only streamlines the procedural process but also makes it feel more intuitive and manageable, enhancing the overall user experience. These findings highlight CAPTAiN’s potential to optimize cognitive workload, thereby improving surgical performance and facilitating more intuitive intraoperative decision-making.

\subsection{Educational and Ethical Context of CAPTAiN}

Hands-on dissection in cadaver laboratories remains a widely regarded benchmark for mastering complex anatomy and acquiring procedural nuance, yet their routine use is restricted by substantial ethical, regulatory, and financial barriers \cite{Bagian2024, Zdilla2022}. Obtaining and stewarding cadavers demands robust consent processes, institutional oversight, specialized facilities, and sustained operating budgets, factors that restrict availability \cite{Champney2015, Champney2018}. Although several jurisdictions have enacted new statutes to encourage donation, establishing the requisite infrastructure is costly and the long-term effect on supply is still uncertain \cite{Brenner2024, DeCaro2021, BoscoloBerto2020, BoscoloBerto2022}.

Within these constraints, VR simulators such as CAPTAiN offer a practical and scalable alternative. The system's demonstrated ability to enhance precision, reduce cognitive load, and support varied trainee skill levels positions it as an accessible adjunct or substitute when cadaveric training is unavailable. Moreover, its use may broaden the reach of surgical education to under-resourced institutions and global regions where traditional training modalities are infeasible.

\subsection{Limitations}
The study’s methodology presents several inherent constraints. First, the spherical modeling of the virtual drill tip—though anatomically representative of diamond bur instruments—imposes geometric simplifications that may compromise edge precision. This approximation risks unintended penetration into critical (red) zones, as the model cannot fully replicate the exact cutting behavior of physical drill bits. This could inadvertently encourage surgeons to drill further into high-risk areas, potentially compromising safety.

Another challenge is the tracking of removed material, which is not commonly addressed in existing systems. While studies have demonstrated promising results in tracking bone material during surgeries \cite{Seim2008, Hemke2020}, integrating such capabilities into the CAPTAiN system would require further investigation. For example, real-time tracking of bone removal could provide additional feedback to surgeons, enhancing their ability to navigate complex anatomical regions.

The substantial preoperative planning required for CAPTAiN is another perceived weakness. While deep learning and statistical methods have shown effectiveness in accurately segmenting bony and soft tissue from CT scans \cite{Seim2008, Hemke2020}, the time and effort required to create the necessary models for navigation may still be comparable to current preoperative planning practices. Further research is needed to assess whether the adoption of CAPTAiN could streamline or enhance this process, potentially reducing the time required for surgeons to approve and adjust plans. For instance, automated segmentation tools and machine learning algorithms could be integrated into the planning pipeline to expedite model generation and improve accuracy.

Navigation in a surgical context also presents unique challenges, particularly when compared to applications like G-code paths used in milling or CNC tasks. Though CAPTAiN allows users to freely navigate within defined boundaries, research has shown that this freedom can sometimes be suboptimal, especially for tasks involving 3-DOF orientation or translation navigation cues \cite{Usevitch2020}. Volume removal tasks, such as those performed in laminectomy, may require different navigation strategies compared to point-to-point tasks, which CAPTAiN does not yet provide.

\subsection{Future Work}
While laminectomy procedures are currently performed without routine use of image-guided surgery (IGS) navigation platforms, the integration of IGS into CAPTAiN presents a transformative opportunity to reduce surgeon strain, standardize training, and improve procedural safety. Existing IGS systems, widely adopted in neurosurgery and orthopedics for tasks such as tumor resection and joint replacement, rely on real-time imaging and spatial tracking to guide surgical tools \cite{Nigam2022, UsevitchDrillingReview2023}. However, these systems have yet to be adapted for laminectomy workflows, which remain heavily dependent on surgeon experience and manual dexterity. By embedding IGS principles into CAPTAiN’s architecture, the system could bridge this gap, offering dynamic visual, haptic, and auditory feedback to assist both trainees and experienced surgeons in navigating complex spinal anatomies.

A critical advancement is to expand CAPTAiN’s capabilities to function as an IGS navigation tool. For instance, integrating intraoperative CT or magnetic resonance imaging (MRI) scans with real-time drill tracking could enable surgeons to visualize bone removal progress relative to preoperative plans, minimizing the risk of over-resection or durotomy. This approach would align with emerging trends in ``smart'' surgical navigation, where AR overlays and force-sensitive instruments enhance precision in bone-cutting tasks \cite{Hoy2017, StrykerMako}. For laminectomy, this could involve color-coded boundaries—similar to those in CAPTAiN’s current training mode—projected directly onto the surgical field, with auditory warnings provided if drilling deviates from the defined trajectory.

The potential of CAPTAiN also extends beyond bone removal, with potential applications in soft tissue procedures and complex surgical navigation. Advancements in soft tissue segmentation and modeling could enable the system to provide guidance during procedures involving delicate structures such as nerves and blood vessels. Moreover, incorporating real-time feedback mechanisms could allow algorithms to assess cut quality and optimize the surgical approach, thereby reducing complications, particularly for less experienced surgeons. These enhancements could also be applied to navigation in anatomically challenging scenarios, such as necrotic bone removal in the femur or hip \cite{Kutzer2011, Sefati2022}, where pre-drilling navigation systems with automated path generation and advanced imaging integration might improve precision and lessen cognitive load.

The integration of CAPTAiN with robotic-assisted surgery platforms represents another promising avenue. By guiding robotic arms and allowing for intraoperative adjustments based on real-time, color-coded feedback, the system could potentially achieve millimeter-level precision. Collectively, these advancements would broaden CAPTAiN’s applicability across both minimally invasive and open procedures and contribute to the development of a unified framework for enhancing surgical precision and patient outcomes through advanced image-guided interventions.

\section{Conclusion}
The findings of this study underscore the potential of CAPTAiN to serve as an effective training tool for laminectomy and other surgical procedures, catering to both novice and experienced practitioners seeking to refine their technical skills. By enabling precise execution of high-priority surgical targets while minimizing deviations into critical regions, the system provides real-time feedback that enhances procedural accuracy. For more seasoned professionals, CAPTAiN may serve as a precision-enhancing adjunct in complex or high-stakes scenarios, whereas for trainees, it offers a structured learning platform that improves technical proficiency and operative confidence. Ultimately, the integration of CAPTAiN into surgical training programs could contribute to higher standards of skill acquisition and improved patient outcomes, warranting further investigation into its long-term impact on surgical education and clinical practice.

\section*{Data Availability}
All processed data (e.g., spine segmentations, analyzed user study results, etc.) supporting the findings of the study are available from the corresponding author upon reasonable request. Raw user study data are not shared to protect participant confidentiality.

\section*{Code Availability}
Upon acceptance of the paper, the underlying code for this study will be made accessible via GitHub. In the meantime, it may be provided to qualified researchers upon reasonable request from the corresponding author.

\section*{Ethics Declaration}
The authors declare no competing interests.

\section*{Acknowledgments}
The authors extend their gratitude to Dr. Craig Forsthoefel and Dr. Christopher Ahuja for their invaluable guidance in the design and preliminary testing of the navigation system. We also thank Miguel Cartagena Reyes for his assistance in preparing the IRB (IRB00343800). The work reported in this paper was supported in part by the National Institutes of Health (NIH) under (a) T32 grant AR067708-07 and (b) R01AR080315, (c) Department of Orthopaedic Surgery at Johns Hopkins University (JHU), (d) a subcontract to JHU from the InnoHK initiative of the Innovation and Technology Commission of the Hong Kong Special Administrative Region Government, in part by JHU internal funds, and (e) JHU 2023 Discovery Grant.

\section*{Copyright}
\textcopyright 2025 IEEE. Personal use of this material is permitted. Permission from IEEE must be obtained for all other uses, in any current or future media, including reprinting/republishing this material for advertising or promotional purposes, creating new collective works, for resale or redistribution to servers or lists, or reuse of any copyrighted component of this work in other works.

The published version of this work is available in \textit{IEEE Transactions on Medical Bionics and Robotics} and can be accessed via the following DOI: 10.1109/TMRB.2025.3589795.

\bibliographystyle{IEEEtran}
\bibliography{references}

\end{document}